%% file: tmlr_submission.tex
\documentclass{article}

\usepackage{microtype}
\usepackage{graphicx}
\usepackage{subcaption}
\usepackage{booktabs} % for professional tables

\usepackage{hyperref}
\usepackage{url}

\usepackage[preprint]{tmlr}

\input{math_commands.tex}

\usepackage{amsmath}
\usepackage{amssymb}
\usepackage{mathtools}
\usepackage{amsthm}
\usepackage{braket}
\usepackage{mathrsfs}
\usepackage{enumerate}
\usepackage{enumitem}
\usepackage{xcolor}

\usepackage[capitalize,noabbrev]{cleveref}

\theoremstyle{plain}
\newtheorem{theorem}{Theorem}[section]

\newtheorem{lemma}[theorem]{Lemma}

\newtheorem{fact}[theorem]{Theorem}

\newtheorem{obs}[theorem]{Observation}
\newtheorem{cor}[theorem]{Corollary}

\theoremstyle{definition}

\theoremstyle{remark}

\providecommand{\E}{\mathbb{E}}

\author{ \name C. Evans Hedges}

\title{Source-Optimal Training is Transfer-Suboptimal}

\begin{document}

\maketitle

\begin{abstract}
We prove that training a source model optimally for its own task is generically suboptimal when the objective is downstream transfer. We study the source-side optimization problem in L2-SP ridge regression and show a fundamental mismatch between the source-optimal and transfer-optimal source regularization: outside of a measure-zero set, $\tau_0^* \neq \tau_S^*$. We characterize the transfer-optimal source penalty $\tau_0^*$ as a function of task alignment and identify an alignment-dependent reversal: with imperfect alignment ($0<\rho<1$), transfer benefits from stronger source regularization, while in super-aligned regimes ($\rho>1$), transfer benefits from weaker regularization. Additionally, in isotropic settings, the decision of whether transfer helps is independent of the target sample size and noise, depending only on task alignment and source characteristics. We verify the linear predictions in a synthetic ridge regression experiment, and we present experiments on MNIST, CIFAR-10, and 20 Newsgroups as evidence that the source-optimal versus transfer-optimal mismatch persists in standard nonlinear transfer learning pipelines.
\end{abstract}

\section{Introduction}

\subsection{Background and Motivation}

A fundamental question in transfer learning is source-side optimization: how should we train the source model if the objective is downstream transfer? We prove a source-optimal versus transfer-optimal mismatch: the source regularization that is optimal for the source task is generically suboptimal for transfer, and outside of a measure-zero set we have $\tau_0^* \neq \tau_S^*$. This mismatch is alignment-dependent: in the standard imperfect-alignment regime ($0<\rho<1$), transfer benefits from stronger source regularization than source-optimal training, while in a super-aligned regime ($\rho>1$) transfer benefits from weaker regularization.

Intuitively, this divergence arises from a geometric mismatch between the source signal and the useful signal for transfer. The source estimator $\hat{\beta}_0$ naturally targets the source vector $w_0$. However, the optimal initialization for the target task is the projection of the target signal $w_1$ onto the source, which corresponds to $\rho w_0$. In the imperfect-alignment regime ($0<\rho<1$), the source estimator is systematically "too large" relative to the target projection; stronger regularization is required to shrink the estimator towards the optimal scale $\rho$. Conversely, in the super-aligned regime ($\rho>1$), the source estimator is "too small" (under-scaled); weaker regularization is preferred to preserve more signal magnitude, even at the cost of higher noise-induced variance. This scaling requirement exists independently of observation noise, though noise further modulates the optimal set-point. Our claims are exact for linear ridge regression; our contribution is isolating a source-side mechanism and making it analytically sharp.

Prior theoretical analyses of transfer learning typically fix the source model and optimize the target-side procedure (or characterize when transfer outperforms training from scratch). In contrast, we study the source-side optimization problem: how should the source model itself be trained to maximize downstream transfer performance? To our knowledge, no prior work derives transfer-optimal source regularization or characterizes its divergence from source-task-optimal training in the general (non-orthonormal) two-task setting we consider.

We begin our exploration through the lens of evaluating when we would expect positive or negative transfer from some source task to a target task. Previous work \cite{dar2022double, lampinen2018analytic, yang2025precise} has shown that the phenomenon of negative transfer is a concern, where the model generated by transfer learning performs worse than if one were to train the model from scratch. Intuition and previous theoretical work \cite{lampinen2018analytic} suggest that if the tasks and input datasets are similar enough, transfer learning techniques should help, but if the target task is sufficiently different, or in the context of continual learning the regime has shifted enough, transfer learning may harm the final model, introducing bias and noise from the initialization.

There has been notable work examining this phase boundary between when one should expect positive or negative transfer. In \cite{dar2022double,dar2024common}, Dar and Baraniuk provided an explicit quantitative phase boundary for freezing/parameter-sharing style transfer in linear models, giving if-and-only-if conditions for when shared representations outperform training from scratch. In \cite{dhifallah2021phase}, Dhifallah and Lu explored the phenomenon in the context of feature freezing. Additionally, feature-overlap driven transitions in linear transfer were explored by Tahir, Ganguli, and Rotskoff in \cite{tahir2024features}. Related phase transitions have been characterized for other linear transfer mechanisms such as hard parameter sharing \cite{yang2025precise}. Complementary asymptotic analyses for fine-tuning from pretrained anchors via gradient descent appear in Ghane, Akhtiamov, and Hassibi \cite{ghane2024universality}, who provide universal asymptotics comparing pretrained versus fine-tuned models under distributional shifts.

The L2-SP (L2-distance to Starting Point) approach was introduced by Li, Grandvalet, and Davoine \cite{xuhong2018explicit}, who showed empirical improvements by penalizing deviation from pre-trained parameters. This approach matches many practical fine-tuning protocols and corresponds to the isotropic elastic weight consolidation (EWC) penalty in \cite{kirkpatrick2017overcoming}. Dar, LeJeune, and Baraniuk \cite{dar2024common} recently analyzed when L2-SP transfer learning outperforms standard ridge regression, assuming task parameters are related by orthonormal transformations and focusing on optimal tuning of the target-task regularization given a fixed source model.

Our work fundamentally differs from these prior analyses by solving the source-side optimization problem. We ask: how should the source model itself be trained to maximize downstream transfer performance? This shift in perspective reveals a new phenomenon: the transfer-optimal source regularization $\tau_0^*$ generically differs from the source-task-optimal choice. We additionally emperically verify that this phenomenon persists across a range of non-linear networks and tasks, providing evidence that this is not an artifact of the linear setting, but instead may be a general phenomenon related to transfer learning pipelines.

\subsection{Setup and Overview of Results}

In this paper we formalize transfer learning from the lens of L2-SP ridge regression. In particular, we have two tasks, Task~0 (the source task) and Task~1 (the target task), as well as corresponding training datasets $(X_0, y_0)$ and $(X_1, y_1)$. We then seek to evaluate the expected out-of-sample risk on Task~1, comparing the ridge/ridgeless solution trained solely on $(X_1, y_1)$ to the L2-SP ridge solution found by first training a ridge/ridgeless model on $(X_0, y_0)$, and then using those model parameters as the prior for ridge/ridgeless training on $(X_1, y_1)$.

Our analysis builds on the foundational work characterizing ridge and ridgeless regression risk in high-dimensional settings \cite{dobriban2018high, hastie2022surprises}, which established the precise asymptotic behavior of these estimators in overparameterized regimes and revealed phenomena such as double descent \cite{belkin2019reconciling} and benign overfitting \cite{bartlett2020benign}. We employ random matrix theory techniques, specifically the deterministic equivalent framework \cite{bai2010spectral,couillet2011random,dobriban2018high,hastie2022surprises}, to derive precise asymptotic characterizations of the estimators' risk and identify sharp phase boundaries for transfer benefit.

Our contributions can be outlined as follows. First, Theorem~\ref{finite-benefit-analysis} provides an if-and-only-if inequality describing when transfer learning will outperform from-scratch training in the finite data, non-isotropic ridge regime. As a corollary, we show that in the finite sample, isotropic, ridgeless case the inequality takes a simple form and is interestingly independent of $n_1$ and $\sigma_1$. This independence is consistent with prior isotropic analyses of freezing-style transfer \cite{dar2022double,lampinen2018analytic}, and our finite-sample isotropic/ridgeless corollary recovers these patterns in the L2-SP setting.

We then examine deterministic equivalents in the asymptotic limit and Theorem~\ref{asymptotic-benefit-analysis} provides a DE characterization and asymptotic boundary for L2-SP ridge transfer with general (non-orthonormal) task vectors. Corollary~\ref{isotropic-asymptotic-benefit} examines the isotropic case, where the decision criterion is determined entirely by whether or not the alignment of the two tasks surpasses a bias and noise term dependent only on Task~0.

From this we arrive at Theorem~\ref{optimal-source-ridge}, which identifies the unique $\tau_0^*$ (source model ridge penalty) that maximizes transfer benefit, given a particular task alignment. Outside of a measure-zero set, $\tau_0^*$ does not coincide with the optimal Task~0 ridge parameter, and notably $\tau_0^*$ is independent of target sample size but depends on task alignment. Together, these results give the first explicit analytical boundary for when Euclidean L2-SP transfer helps with general (non-orthonormal) task vectors in overparameterized ridge models, including general covariance and DE limits, and provide surprising insights into optimally training source models for the purpose of transfer learning.

Finally, we establish in Corollary~\ref{snr-threshold-cor} an alignment-dependent phase transition for the optimal source penalty. Under standard imperfect task alignment conditions, $\tau_0^*$ is always strictly greater than the source-optimal ridge penalty. However, in super-aligned regimes this relationship reverses. This implies that maximizing transfer benefit requires adjusting regularization based on the geometric relationship between tasks, rather than optimizing for source performance alone.

Though derived in linear models, these results isolate a core mechanism that may persist in overparameterized nonlinear networks. Additionally although standard SGD fine-tuning protocols do not necessarily use an L2-SP penalty, this mechanism models the implicit regularization of SGD fine-tuning where the optimization trajectory remains anchored in the basin of the initialization. In Section~\ref{sec:experiments} we validate the theory-derived phase transition in a synthetic ridge setting and then use standard transfer learning experiments on MNIST, CIFAR-10, and 20 Newsgroups to probe how source-optimal versus transfer-optimal regularization behaves in nonlinear networks, finding a consistent preference for over-regularization for transfer across all tested domains.

These results challenge the conventional practice of optimizing source models solely for their own performance. For practitioners training foundation models intended for transfer, our analysis suggests that regularization strategies should explicitly account for the downstream transfer objective and source data quality.

The remainder of the paper is outlined as follows. Section~\ref{sec:prelim} sets up the model and assumptions. Section~\ref{sec:main} presents the core mathematical results. Section~\ref{sec:experiments} provides empirical validation of our theoretical predictions on synthetic data, MNIST, CIFAR-10, and 20 Newsgroups. We conclude with Section~\ref{sec:discussion} and discuss implications and limitations; proofs are deferred to the Appendix.

\section{Preliminaries}\label{sec:prelim}

We let $Z_0, Z_1$ be $n_0 , n_1 \times p$ random matrices with entries taken iid with mean $0$, variance $1$, and finite $2+\epsilon$ moments and Lindeberg condition (see \cite{bai2010spectral} for more information). These assumptions are sufficient for the deterministic equivalents we will examine later, but it is safe also to simplify these assumptions taking entries in $Z_i$ to be iid $N(0, 1)$. We additionally assume we are in the overparameterized regime, with $n_0, n_1 < p-1$. We let $\Sigma_0, \Sigma_1$ be covariance matrices and $X_i = Z_i \Sigma_i^{1/2}$. Next we have true signal vectors $w_0, w_1$ and let $y_i = X_i w_i + \epsilon_i$ where $\epsilon_i \sim N(0, \sigma_i^2 I)$. We assume the signal norms $||w_0||, ||w_1||$ remain bounded (i.e., $O(1)$) as $p \to \infty$, ensuring that the signal and noise contributions to the risk remain comparable in the asymptotic limit.

We will frequently quantify task relatedness by the normalized alignment
$$\rho := \frac{\braket{w_0, w_1}}{||w_0||^2}.$$
We refer to $0<\rho<1$ as the \emph{imperfect alignment} regime and $\rho>1$ as the \emph{super-aligned} regime.

We adopt the L2-SP approach found in \cite{xuhong2018explicit} and examine the following estimators:
\begin{align*}
\hat{\beta}_{0}(\lambda_0)
    &:= \arg\min_{\beta} \; \|y_0 - X_0 \beta\|^2 + \lambda_0 \|\beta\|^2 \\
     &= (X_0^\top X_0 + \lambda_0 I)^{-1} X_0^\top y_0, \\[0.8em]
\hat{\beta}_{1}^{\mathrm{S}}(\lambda_1)
    &:= \arg\min_{\beta} \; \|y_1 - X_1 \beta\|^2 + \lambda_1 \|\beta\|^2 \\
     &= (X_1^\top X_1 + \lambda_1 I)^{-1} X_1^\top y_1, \\[0.8em]
\hat{\beta}_{1}^{\mathrm{TL}}(\lambda_1 \mid \hat{\beta}_0)
    &:= \arg\min_{\beta} \; \|y_1 - X_1 \beta\|^2 + \lambda_1 \|\beta - \hat{\beta}_0(\lambda_0)\|^2 \\
    &= (X_1^\top X_1 + \lambda_1 I)^{-1} \big(X_1^\top y_1 + \lambda_1 \, \hat{\beta}_0(\lambda_0)\big).
\end{align*}
We also will take the notation that $\hat{\beta}_{0}(0) = \lim_{\lambda_0 \searrow 0} \hat{\beta}_{0}(\lambda_0)$ represents the ridgeless estimator (and similarly for $\hat{\beta}_1^{S}$ and $\hat{\beta}_1^{TL}$). In this setting of course $\hat{\beta}_0$ represents our ridge/ridgeless estimator for Task~0, $\hat{\beta}_1^{S}$ is our standard ridge/ridgeless estimator for Task~1, and $\hat{\beta}_1^{TL}$ is our transfer learning estimator for Task~1 that takes the solution of Task~0 as its prior.

We note here that $\hat{\beta}_{1}^{\mathrm{TL}}(\lambda_1 \mid \hat{\beta}_0)$ corresponds exactly to the MAP estimator with Gaussian prior $\beta \sim N(\hat{\beta}_0, \lambda_1^{-1} I)$ and matches practical L2-SP fine-tuning \cite{xuhong2018explicit} and the isotropic EWC penalty \cite{kirkpatrick2017overcoming}. While our theoretical analysis assumes an explicit L2-SP penalty, standard fine-tuning protocols rely on the implicit regularization of SGD initialized at $\theta_0$. For limited training horizons, the optimization trajectory remains anchored in the basin of $\theta_0$, functionally approximating the L2-SP constraint. Thus, we expect the regularization-variance trade-offs identified in our ridge analysis to persist in standard fine-tuning. We also note that we do not reweight our penalty by a task metric $H$ and stick with standard Euclidean distancing for our ridge penalty. This differs from the whitened/metric-based formulations and also more accurately reflects typical implementations of L2-SP and fine-tuning where the penalty is applied in Euclidean parameter space without whitening.

To evaluate out-of-sample performance of these estimators we will use their expected prediction risk:
\[
\mathrm{Risk}_1(\beta) := \mathbb{E}_{x_1\sim N(0,\Sigma_1)}[(x_1^\top\beta - x_1^\top w_1)^2] = \|\beta - w_1\|_{\Sigma_1}^2 ,
\]
where $||v||_{\Sigma}^2 = v^\top \Sigma v $.

Our first goal is to understand when L2-SP transfer improves expected Task~1 risk, namely when
\[
\mathrm{Risk}_1(\hat{\beta}_{1}^{\mathrm{S}}(\lambda_1)) > \mathrm{Risk}_1(\hat{\beta}_{1}^{\mathrm{TL}}(\lambda_1 \mid \hat{\beta}_0(\lambda_0))) .
\]

We will additionally make use of the Frobenius norm in $\Sigma$ geometry, which we will denote as follows:
$$||A||_{\Sigma, F}^2 = \operatorname{Tr} \left( A^\top \Sigma A\right). $$

Finally, some common matrices we will be using deserve their own notation, and we define that here: let
$$M_{\lambda_1}^{(i)} = \left(X_i^\top X_i + \lambda_1 I\right)^{-1}$$
and note that
$$M_{\lambda_1}^{(1)}X_1^\top X_1 - I = -\lambda_1 M_{\lambda_1}^{(1)}.$$
Additionally let
$$P_i = X_i^+ X_i,$$
where $X_i^+$ is the Moore-Penrose pseudoinverse, so that $P_i$ is the orthogonal projector in $\mathbb{R}^p$ onto $\mathrm{row}(X_i)$ and $I-P_i$ projects onto $\ker(X_i)$.

\section{Mathematical Results}\label{sec:main}

\subsection{Finite Sample Risk Formulas}

First we recall the expected out of sample risk of ridge regression (see \cite{hastie2022surprises} for a modern reference)
\begin{obs} The expected risk of ridge regression is
$$R^{S}(\lambda_1) =\lambda_1^2  \mathbb{E} \left[ || M_{\lambda_1}^{(1)} w_1  ||_{\Sigma_1}^2 \right] + \sigma_1^2 \mathbb{E} \left[ || M_{\lambda_1}^{(1)} X_1^\top||_{\Sigma_1, F}^2\right]. $$
\end{obs}

Next we compute the finite sample risk for L2-SP ridge regression:
\begin{lemma}\label{transfer-risk} The expected risk of the transfer estimator decomposes into pure bias, variance induced by the $\beta_0$ prior, and variance induced by estimation error:
$$R^{TL}(\lambda_1) = B^{TL}(\lambda_1) + \sigma_0^2 V_0^{TL}(\lambda_1) + \sigma_1^2 V_1^{TL}(\lambda_1)$$
with:
$$B^{TL}(\lambda_1) = \lambda_1^2  \E \left[ ||  M_{\lambda_1}^{(1)}M_{\lambda_0}^{(0)}X_0^\top X_0 w_0 - M_{\lambda_1}^{(1)} w_1||_{\Sigma_1}^2 \right]$$
$$V_0^{TL}(\lambda_1) =  \lambda_1^2 \E \left[ || M_{\lambda_1}^{(1)}M_{\lambda_0}^{(0)}X_0^\top||_{\Sigma_1, F}^2  \right]$$
$$V_1^{TL}(\lambda_1) =  \E \left[|| M_{\lambda_1}^{(1)} X_1^\top ||_{\Sigma_1, F}^2 \right] .$$
\end{lemma}

The following theorem characterizes when transfer helps by comparing the bias introduced by starting from zero (standard ridge) versus starting from the source parameters (L2-SP). Transfer succeeds when the alignment between the tasks, after appropriate filtering through the ridge resolvent and covariance geometry, exceeds a threshold determined by source bias and noise. Note that since the Task~1 variance portion of risk for both of these estimators is the same, it is independent from the decision of whether or not transfer will benefit.

\begin{fact}\label{finite-benefit-analysis} In the finite sample case with $\lambda_1 > 0$, we gain benefit from transfer learning ($R^{TL}(\lambda_1) < R^S(\lambda_1)$) if and only if:
\begin{align*}
&2 \E \left[ \braket{ M_{\lambda_1}^{(1)}M_{\lambda_0}^{(0)}X_0^\top X_0 w_0 ,  M_{\lambda_1}^{(1)} w_1}_{\Sigma_1}\right] \\
&\quad > \E \left[ ||  M_{\lambda_1}^{(1)}M_{\lambda_0}^{(0)}X_0^\top X_0 w_0 ||_{\Sigma_1}^2 \right]  + \sigma_0^2 \E\left[ || M_{\lambda_1}^{(1)} M_{\lambda_0}^{(0)} X_0^\top ||_{\Sigma_1, F}^2 \right]. 
\end{align*}
\end{fact}

As a corollary, in the ridgeless limit we consider when $X_i$ is taken to be isotropic and Gaussian so that the Wishart formulas apply exactly:

\begin{cor}\label{finite-ridgeless-boundary} In the finite case, if $\Sigma_i = I$ and $\lambda_1 = \lambda_0 = 0$ and $X_i$ is taken to be Gaussian, then $R^{TL}(0) < R^{S}(0)$ if and only if:
$$ 2 \braket{w_0, w_1} > ||w_0||^2 + \sigma_0^2 \frac{p}{p-n_0-1}. $$
\end{cor}

A noteable feature of the isotropic ridgeless boundary is its complete independence from $n_1$ and $\sigma_1$: the transfer decision depends only on task alignment $\braket{w_0, w_1}$ and source characteristics $(||w_0||, \sigma_0, n_0, p)$. If transfer outperforms training from scratch with 10 target samples, it also outperforms training from scratch with 10{,}000 target samples (and the same statement holds across target noise levels). Collecting more target data does not change whether transfer helps in this isotropic regime; only task alignment and source-side quality matter.

We also observe here that the transfer region monotonically shrinks as $\sigma_0$ grows. Thus, in order to maximize the potential transfer benefit it is essential to ensure source task noise is as small as possible.

\subsection{Deterministic Equivalents and Asymptotics}

We will now examine asymptotics for the ridge phase transition identified above and use deterministic equivalents to understand the limiting phase transition. To establish the core deterministic equivalents, we will scale our $\lambda_i$ with $n_i$ and let $\tau_i =  \lambda_i / n_i$. We define the following ridge resolvent:
Let $S_i := n_i^{-1} X_i^\top X_i$ and $\gamma_i := \lim_{p\to\infty} p/n_i$. Under standard Bai--Silverstein assumptions (e.g.\ bounded spectral norm of $\Sigma_i$ and weak convergence of its empirical spectral distribution), the resolvent $(S_i+\tau_i I)^{-1}$ admits a deterministic equivalent of the form
$$Q_i(\tau_i) = \Big(\tau_i I + \tilde{\delta}_i(\tau_i)\Sigma_i\Big)^{-1},$$
where the scalar pair $(\delta_i(\tau_i),\tilde{\delta}_i(\tau_i))$ is the unique positive solution to
$$\delta_i(\tau_i) = \frac{1}{n_i}\operatorname{Tr}\big(\Sigma_i Q_i(\tau_i)\big), \qquad \tilde{\delta}_i(\tau_i)=\frac{1}{1+\delta_i(\tau_i)}.$$

We will use the notation $A_n \asymp B_n$ to denote deterministic equivalent convergence, meaning that for any sequence of deterministic matrices $D_n$ with uniformly bounded spectral norm, we have $\operatorname{Tr}(D_n(A_n - B_n)) \to 0$ almost surely as $n \to \infty$. This is the standard notion of weak deterministic equivalence used in random matrix theory \cite{bai2010spectral}.

First, the following deterministic equivalents will be useful:

\begin{obs}\label{bai-silverstein-de} Under standard Bai-Silverstein conditions (\cite{bai2010spectral}), as $p, n_i \rightarrow \infty$ with $p/n_i \rightarrow \gamma_i$:
$$n_1 M_{\lambda_1}^{(1)} = (S_1+\tau_1 I)^{-1} \asymp Q_1(\tau_1),$$
$$ \lambda_1 M_{\lambda_1}^{(1)} \asymp \tau_1 Q_1(\tau_1),$$
and
$$n_0 M_{\lambda_0}^{(0)} X_0^\top X_0 M_{\lambda_0}^{(0)} \asymp Q_0(\tau_0) - \tau_0 Q_0(\tau_0)^2.$$
\end{obs}

Finally, define $t(\tau_0, \tau_1)$ by: 
$$ \lim_{p \rightarrow \infty} p^{-1} \operatorname{Tr}\left( Q_1(\tau_1) \Sigma_1 Q_1(\tau_1) \left( Q_0(\tau_0) - \tau_0 Q_0(\tau_0)^2 \right)\right), $$
Under the Bai-Silverstein assumptions, the eigenvalue distributions of $\Sigma_0$ and $\Sigma_1$ converge weakly and the resolvents $Q_i(\tau_i)$ are uniformly bounded, ensuring that this limit exists and equals the limit of the corresponding trace per dimension. By independence of $X_0$ and $X_1$, we may substitute these deterministic equivalents into the results from Theorem~\ref{finite-benefit-analysis} to arrive at the following asymptotic decision criterion:

\begin{fact}\label{asymptotic-benefit-analysis} In the asymptotic limit, we gain benefit from transfer learning ($R^{TL}(\lambda_1) < R^S(\lambda_1)$) if and only if:
\begin{align*}
&2 \braket{ Q_1(\tau_1) (I - \tau_0 Q_0(\tau_0)) w_0,  Q_1(\tau_1) w_1}_{\Sigma_1} \\
&\quad >  ||  Q_1(\tau_1) (I - \tau_0 Q_0(\tau_0)) w_0 ||_{\Sigma_1}^2  +  \sigma_0^2 \gamma_0 t(\tau_0, \tau_1). 
\end{align*}
\end{fact}

And in the isotropic case:
\begin{cor}\label{isotropic-asymptotic-benefit} In the isotropic case where $\Sigma_0 = \Sigma_1 = I$, we (asymptotically) gain benefit from transfer learning ($R^{TL}(\lambda_1) < R^S(\lambda_1)$) if and only if:
$$ 2 \braket{w_0, w_1} >  (1-\tau_0 a_0) ||w_0||^2 + \sigma_0^2 \gamma_0 a_0, $$
where $a_0$ is the unique positive solution to
$$\tau_0 \gamma_0 a_0^2 + (\tau_0 + 1 - \gamma_0)a_0 - 1 = 0.$$
\end{cor}

As in the finite sample isotropic ridgeless case, in the isotropic DE (ridge) setting this decision boundary is independent of $\tau_1$, $\gamma_1$, and $\sigma_1$. In particular, whether transfer helps is a source-side question: it does not depend on target sample size, target noise, or target regularization. 

As in the finite sample isotropic ridgeless case, alignment of the two tasks must pass a threshold that is independent of other target task characeristics. This transfer benefit region is again monotonically shrinking in $\sigma_0$, and may be maximized by minimizing the quantity on the right. By fixing $\gamma_0$ we arrive at the following result:

\begin{fact}\label{optimal-source-ridge} In the asymptotic setting with isotropic data and fixed source model overparameterization $\gamma_0$, for any fixed normalized alignment $\rho = \braket{w_0, w_1}/||w_0||^2$, there exists a unique source ridge penalty $\tau_0^*$ that maximizes the transfer benefit $\Delta R = R^S - R^{TL}$. Additionally, whenever $\braket{w_0, w_1} \neq ||w_0||^2$, the optimal $\tau_0^*$ for transfer learning differs from the optimal ridge penalty for Task~0 performance.
\end{fact}

As a corollary, we identify that there is a consistent relationship between transfer-optimal and source-optimal regularization.

\begin{cor}\label{snr-threshold-cor} Let $\tau_0^*$ be the transfer-optimal regularization penalty and $\tau_S^*$ be the source-optimal regularization penalty.
\begin{itemize}
    \item If the tasks are imperfectly aligned ($0<\rho<1$), then $\tau_0^* > \tau_S^*$ (transfer requires stronger regularization).
    \item If the tasks are super-aligned ($\rho>1$), then $\tau_0^* < \tau_S^*$ (transfer requires weaker regularization).
\end{itemize}
This phase transition depends solely on task alignment and holds for all noise levels $\sigma_0 > 0$.
\end{cor}

In particular, this shows that when the tasks are imperfectly aligned ($0<\rho<1$), the transfer-optimal solution always uses more regularization than one would typically use for the source task alone.

\section{Empirical Validation}\label{sec:experiments}

In this section we provide a controlled synthetic experiment to validate the precise phase transition predictions. Additionally, while our theoretical analysis focuses on linear ridge regression, we study nonlinear networks on MNIST, CIFAR-10, and 20 Newsgroups to test whether the source-optimal versus transfer-optimal misalignment persists beyond linear models. 

\subsection{Synthetic Validation of Phase Transition}

\subsubsection{Experimental Setup}
To verify the alignment-dependent phase transition predicted by Corollary~\ref{snr-threshold-cor}, we conduct a controlled synthetic experiment using the generative model defined in Section~\ref{sec:prelim}. We set $p=500$, $n_{source}=250$ ($\gamma=2.0$), and $n_{target}=50$. We fix the source noise $\sigma_0^2=1.0$ and target noise $\sigma_1^2=0.1$.

We sweep the task alignment $\rho = \braket{w_0, w_1} / ||w_0||^2$ from $0.5$ to $1.5$, covering both the imperfect alignment regime ($\rho < 1$) and the super-aligned regime ($\rho > 1$). For each alignment level, we:
1. Generate source and target data according to the specified alignment.
2. Train source ridge models over a logarithmic grid of regularization strengths $\lambda$.
3. Select the source-optimal $\lambda_S^*$ that minimizes risk on the source task.
4. For each trained source model, fine-tune on the target task (using L2-SP with fixed target regularization $\lambda_1=0.1$) and identify the transfer-optimal source regularization $\lambda_{TL}^*$.
5. Compute the ratio $\lambda_{TL}^* / \lambda_S^*$.

We repeat this procedure across 10 random seeds to estimate confidence intervals.

\subsubsection{Results}
\begin{figure}[t]
\centering
\includegraphics[width=0.65\textwidth]{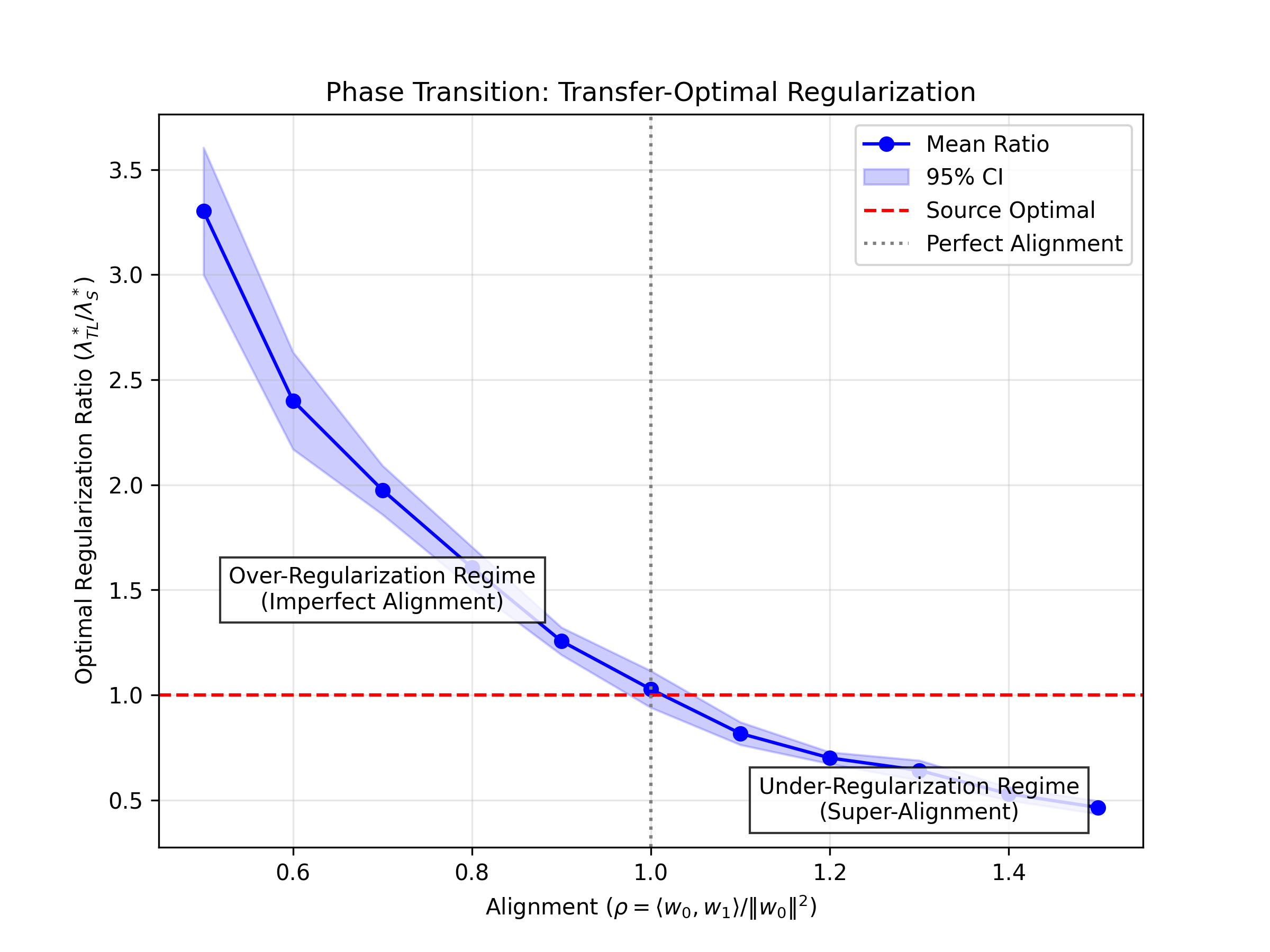}
\caption{Synthetic validation of the alignment-dependent phase transition. The y-axis shows the ratio of transfer-optimal to source-optimal regularization ($\lambda_{TL}^*/\lambda_S^*$). A ratio $>1$ indicates over-regularization is beneficial (Standard Regime), while $<1$ indicates under-regularization is optimal (Super-Aligned Regime). The shaded region represents the 95\% confidence interval over 10 seeds.}
\label{fig:phase_transition}
\end{figure}

Figure~\ref{fig:phase_transition} displays the ratio of transfer-optimal to source-optimal regularization as a function of task alignment. The results match our theoretical predictions:
\begin{itemize}
\item \textbf{Over-Regularization Regime ($\rho < 1$):} When alignment is imperfect (standard transfer), the ratio is consistently $>1$, indicating that $\lambda_{TL}^* > \lambda_S^*$. For example, at $\rho=0.8$, the transfer-optimal penalty is approximately $1.6\times$ larger than the source-optimal penalty.
\item \textbf{Under-Regularization Regime ($\rho > 1$):} When the tasks are super-aligned, the ratio drops below $1$, confirming that $\lambda_{TL}^* < \lambda_S^*$. At $\rho=1.2$, the optimal source penalty for transfer is roughly $0.7\times$ the source-optimal value.
\item \textbf{Phase Transition:} The crossover occurs precisely at $\rho=1.0$, matching the theoretical boundary where the target signal magnitude equals the source signal magnitude.
\end{itemize}

\subsection{Nonlinear Transfer Learning Experiments}

To test whether the source-optimal versus transfer-optimal mismatch persists beyond linear models, we conduct standard transfer learning experiments across three domains: vision (MNIST, CIFAR-10) and text (20 Newsgroups). Each experiment follows the same protocol: (1) train source models with varying weight decay, (2) transfer to target task with fixed fine-tuning protocol, (3) compare source-optimal versus transfer-optimal source regularization.

\subsubsection{Experimental Setup}

\textbf{MNIST.} We use a 2-layer MLP (784$\to$128$\to$64$\to$5) and split digits into source (round digits: 0,3,6,8,9) and target (angular digits: 1,2,4,5,7). The target training set is subsampled to 10\% to create a realistic limited-data transfer scenario.

\textbf{CIFAR-10.} We use a small CNN (two conv layers, one FC layer, $\sim$530K parameters) and split classes into source (animals: bird, cat, deer, dog, frog) and target (vehicles+: airplane, automobile, horse, ship, truck). The target training set is subsampled to 5\%.

\textbf{20 Newsgroups.} We use TF-IDF features (5000 dimensions) with a 2-layer MLP (5000$\to$256$\to$128$\to$num\_classes) and split categories into source (tech: comp.*, sci.*) and target (non-tech: rec.*, talk.*, misc.*, alt.atheism). The target training set is subsampled to 10\%.

For all experiments, we sweep source weight decay over $[0, 10^{-5}, 5{\times}10^{-5}, 10^{-4}, 5{\times}10^{-4}, 10^{-3}, 5{\times}10^{-3}, 10^{-2}]$ and use a fixed transfer protocol (learning rate $10^{-3}$, weight decay $10^{-4}$). Results are averaged over 5 random seeds.

\subsubsection{Results}

\begin{figure}[t]
\centering
\includegraphics[width=1\textwidth]{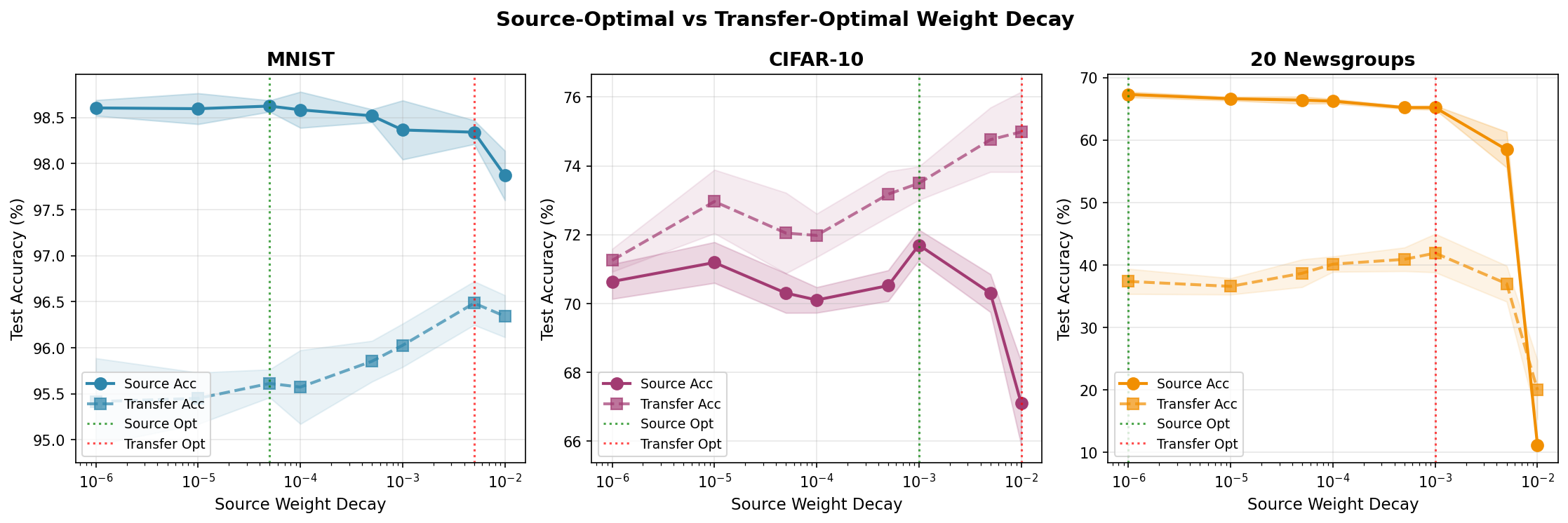}
\caption{Source-optimal versus transfer-optimal weight decay across three standard transfer learning benchmarks. Each panel shows source accuracy (solid) and transfer accuracy (dashed) as a function of source weight decay, with vertical lines marking optimal values. Across all domains, transfer-optimal performance occurs at stronger source regularization than source-optimal performance, consistent with the over-regularization regime for imperfectly aligned tasks.}
\label{fig:nonlinear_transfer}
\end{figure}

Figure~\ref{fig:nonlinear_transfer} shows that across all three domains, the source-optimal and transfer-optimal regularization differ substantially. In each case, transfer-optimal performance (red dashed line) occurs at stronger source regularization than source-optimal performance (green dashed line), consistent with the over-regularization regime predicted by our theory for imperfectly aligned tasks.

\textbf{MNIST:} Source accuracy peaks at low weight decay ($\sim$$10^{-5}$), while transfer accuracy peaks at higher weight decay ($\sim$$10^{-2}$), a difference of approximately three orders of magnitude.

\textbf{CIFAR-10:} Source accuracy peaks around $10^{-3}$, while transfer accuracy peaks at $10^{-2}$, demonstrating the mismatch persists in convolutional architectures.

\textbf{20 Newsgroups:} Source accuracy peaks at low weight decay ($\sim$$10^{-5}$), while transfer accuracy peaks around $10^{-3}$, showing the phenomenon extends to text classification with TF-IDF features.

These results consistently validate that source-optimal training is suboptimal for transfer learning across vision and NLP domains, with transfer uniformly preferring stronger source regularization.

\section{Conclusion}\label{sec:discussion}

We have proven a fundamental misalignment in transfer learning: training a source model to minimize its own risk is generically suboptimal for maximizing transfer benefit. This source-optimal versus transfer-optimal divergence implies that source regularization should be chosen with the downstream objective in mind and depends on task alignment. We provide explicit transfer-versus-scratch boundaries for L2-SP ridge that hold at finite $p,n$ and extend to general covariance via deterministic equivalents. In isotropic limits, whether transfer helps is independent of target sample size and noise: if transfer helps with 10 target samples, it helps with 10{,}000. Optimizing the source model target task performance yields a unique source ridge $\tau_0^*$ that differs generically from the source-task-optimal choice.

These results give an analytical foundation for the idea that pretraining optimized for transfer differs fundamentally from pretraining aimed at standalone accuracy. They suggest re-evaluating regularization and objective design in pretraining, foundation-model, and continual-learning pipelines. In particular, practitioners training source models with the goal of downstream transfer should carefully consider the alignment between source and target tasks. In standard imperfect alignment scenarios, stronger regularization is required to shrink the source estimator towards the optimal target projection.

Our experiments on MNIST, CIFAR-10, and 20 Newsgroups indicate that the misalignment between source-optimal and transfer-optimal regularization persists beyond the linear setting across vision and text domains. In all nonlinear regimes we tested, the transfer-optimal choice consistently involves stronger source regularization than source-optimal training. We did not observe a super-aligned regime where under-regularizing the source improves transfer, suggesting that exhibiting the under-regularization phase may be substantially more difficult in standard deep-learning pipelines where feature learning becomes an important aspect of the model training process.

\subsection{Practical Implications}

Our results provide several concrete insights for practitioners working with transfer learning systems. First, minimizing source noise is critical for maximizing the region of positive transfer, as the transfer benefit shrinks monotonically with $\sigma_0^2$. This suggests that investing in data cleaning, noise reduction, or improved labeling for the source task can pay significant dividends for downstream transfer, beyond simply improving source task performance. Second, the independence of the transfer decision from target sample size and noise in isotropic settings suggests that if transfer is beneficial for one target dataset size, it remains beneficial across different scales of target data, provided the task structure remains similar. This can simplify model selection in scenarios where target data availability is uncertain or variable.

The alignment-dependent phase transition provides actionable guidance for model training in linear ridge: when tasks exhibit standard imperfect alignment ($\braket{w_0, w_1} < ||w_0||^2$), one should increase regularization beyond what would be optimal for source task performance alone, effectively trading off some source performance for better transferability. In the nonlinear regimes we tested across MNIST, CIFAR-10, and 20 Newsgroups, we consistently observed this over-regularization behavior for transfer. 

Since most practical purposes likely lie in the standard imperfect alignment regime, it is likely the case that over-regularization during source training may lead to better outcomes for transfer learning. Additionally, as noted in Figure~\ref{fig:phase_transition}, this divergence can be quite non-trivial, where even with a relatively strong alignment of $0.8$, the transfer-optimal regularization penalty was $\sim60$\% higher than source-optimal.

We note here that while our analysis focuses on L2-SP ridge regression, our experiments across MNIST, CIFAR-10, and 20 Newsgroups indicate these insights extend qualitatively to more complex fine-tuning protocols in deep learning, suggesting that foundation model training pipelines may benefit from regularization strategies that explicitly account for downstream transfer objectives rather than solely optimizing upstream performance.

\subsection{Scope and Future Directions}

Our results establish the source-optimal versus transfer-optimal phenomenon in the tractable linear setting and provide a foundation for understanding transfer in complex models. The precise functional forms and boundaries we derive are specific to linear ridge regression. Our experiments on MNIST, CIFAR-10, and 20 Newsgroups suggest the source/transfer regularization misalignment persists in nonlinear networks across vision and text domains, while the super-aligned under-regularization regime appears difficult to realize in standard deep-learning setups. We assume Gaussian noise and employ either exact isotropy or Bai-Silverstein conditions for our deterministic equivalent results. Extensions to more general noise models and covariance structures remain important open problems.

We focus on the L2-SP protocol, which penalizes Euclidean distance from the source parameters. This differs from more sophisticated fine-tuning methods that might incorporate task-specific metrics, feature-space regularization, or adaptive penalties based on parameter importance. Additionally, L2-SP represents a closed-form ridge solution, while practical deep learning relies on iterative optimization via gradient descent. However, the implicit regularization of early-stopped SGD typically keeps parameters close to initialization, qualitatively matching the L2-SP behavior. The recent work of Ghane, Akhtiamov, and Hassibi \cite{ghane2024universality} on gradient-descent-based fine-tuning provides complementary insights for the iterative case, and connecting these perspectives remains an interesting direction.

Finally, our results focus on two-task transfer, whereas many practical scenarios involve continual learning across multiple sequential tasks or multi-task learning with many simultaneous objectives. The interplay between multiple tasks and optimal source regularization represents a rich area for future investigation.

Connecting our ridge-based findings to gradient-descent dynamics in deep learning represents an important bridge to practice. Understanding how implicit regularization in over-parameterized neural networks interacts with the mechanisms we identify could inform training procedures for foundation models. In particular, investigating whether early stopping, learning rate schedules, or architectural choices can implicitly achieve transfer-optimal regularization even when explicit penalties are not used would provide actionable insights for practitioners. Analyzing multi-task and continual learning scenarios where multiple target tasks must be considered simultaneously could reveal how to optimize source training when facing diverse downstream objectives with potentially conflicting requirements.

Though derived in the tractable setting of linear ridge regression, our results isolate a fundamental tension between optimizing for source performance versus transfer capability that likely manifests across a broad range of learning systems. By making this tension analytically precise, we hope to inform the design of pretraining objectives that explicitly account for downstream transfer goals.

% \section*{Impact Statement}
% This paper presents theoretical work advancing the understanding of transfer learning and regularization. There are many potential societal consequences of improved transfer learning (both positive, like better medical diagnostics from limited data, and negative, like dual-use capabilities), none which we feel must be specifically highlighted here beyond standard considerations in the field.

\bibliography{mybib}
\bibliographystyle{tmlr}

%%%%%%%%%%%%%%%%%%%%%%%%%%%%%%%%%%%%%%%%%%%%%%%%%%%%%%%%%%%%%%%%%%%%%%%%%%%%%%%
%%%%%%%%%%%%%%%%%%%%%%%%%%%%%%%%%%%%%%%%%%%%%%%%%%%%%%%%%%%%%%%%%%%%%%%%%%%%%%%
% APPENDIX
%%%%%%%%%%%%%%%%%%%%%%%%%%%%%%%%%%%%%%%%%%%%%%%%%%%%%%%%%%%%%%%%%%%%%%%%%%%%%%%
%%%%%%%%%%%%%%%%%%%%%%%%%%%%%%%%%%%%%%%%%%%%%%%%%%%%%%%%%%%%%%%%%%%%%%%%%%%%%%%
\newpage
\appendix
\onecolumn

\section{Detailed Proofs}

Here we will re-state and prove the relevant Lemmas and Theorems mentioned above.

\subsection{Finite Sample Risk Formulas} We begin by re-writing our L2-SP transfer learning estimator for the reader's convenience:
$$\hat{\beta}_{1}^{\mathrm{TL}}(\lambda_1 \mid \hat{\beta}_0) = (X_1^\top X_1 + \lambda_1 I)^{-1}\Big(X_1^\top y_1 + \lambda_1 \, \hat{\beta}_0(\lambda_0)\Big). $$

We additionally note that expected risk can be computed by:
$$R(\beta) = E\left[ ||\beta - w_1||_{\Sigma_1}^2 \right].$$

We can now prove the first important Lemma:

\textbf{Lemma \ref{transfer-risk}} The expected risk of the transfer estimator decomposes into pure bias, variance induced by the $\beta_0$ prior, and variance induced by estimation error:
$$R^{TL}(\lambda_1) = B^{TL}(\lambda_1) + \sigma_0^2 V_0^{TL}(\lambda_1) + \sigma_1^2 V_1^{TL}(\lambda_1)$$
with:
$$B^{TL}(\lambda_1) = \lambda_1^2  \E \left[ ||  M_{\lambda_1}^{(1)}M_{\lambda_0}^{(0)}X_0^\top X_0 w_0 - M_{\lambda_1}^{(1)} w_1||_{\Sigma_1}^2 \right]$$
$$V_0^{TL}(\lambda_1) =  \lambda_1^2 \E \left[ || M_{\lambda_1}^{(1)}M_{\lambda_0}^{(0)}X_0^\top||_{\Sigma_1, F}^2  \right]$$
$$V_1^{TL}(\lambda_1) =  \E \left[|| M_{\lambda_1}^{(1)} X_1^\top ||_{\Sigma_1, F}^2 \right] .$$

\begin{proof} We verify the decomposition as follows. First note:
$$\hat{\beta}_{1}^{\mathrm{TL}}(\lambda \mid \hat{\beta}_0 )- w_1 = \lambda M_\lambda^{(1)} \hat{\beta}_0^S(\lambda_0) + \hat{\beta}_1^S(\lambda) - w_1$$
$$= \lambda M_\lambda^{(1)}M_{\lambda_0}^{(0)} X_0^\top y_0 + M_\lambda^{(1)} X_1^\top y_1 - w_1$$
$$= \lambda M_\lambda^{(1)}M_{\lambda_0}^{(0)} X_0^\top X_0 w_0 + \lambda M_\lambda^{(1)}M_{\lambda_0}^{(0)} X_0^\top \epsilon_0 +  M_\lambda^{(1)} X_1^\top X_1 w_1 +  M_\lambda^{(1)} X_1^\top \epsilon_1 - w_1 $$
$$=  \left(\lambda M_\lambda^{(1)}M_{\lambda_0}^{(0)} X_0^\top X_0 + \left( M_\lambda^{(1)} X_1^\top X_1 - I \right) \right)w_1 + \left(\lambda M_\lambda^{(1)}M_{\lambda_0}^{(0)} X_0^\top \right) \epsilon_0 + \left( M_\lambda^{(1)} X_1^\top \right)\epsilon_1 .$$
We then take expectation of the squared $\Sigma_1$ norm of the above expression and since $\epsilon_0, \epsilon_1$ are taken independently with mean $0$, the cross terms cancel and we are left with the desired result.
\end{proof}

We now remind the reader that the expected risk of the standard Ridge estimator is
$$R^{S}(\lambda_1) =\lambda_1^2  \mathbb{E} \left[ || M_{\lambda_1}^{(1)} w_1  ||_{\Sigma_1}^2 \right] + \sigma_1^2 \mathbb{E} \left[ || M_{\lambda_1}^{(1)} X_1^\top||_{\Sigma_1, F}^2\right]. $$

We can therefore compare these quantities to identify when we expect Transfer to outperform training from scratch.

\textbf{Theorem \ref{finite-benefit-analysis}}. In the finite sample case with $\lambda_1 > 0$, we gain benefit from transfer learning ($R^{TL}(\lambda_1) < R^S(\lambda_1)$) if and only if:
$$2 \E \left[ \braket{ M_{\lambda_1}^{(1)}M_{\lambda_0}^{(0)}X_0^\top X_0 w_0 ,  M_{\lambda_1}^{(1)} w_1}_{\Sigma_1}\right] > \E \left[ ||  M_{\lambda_1}^{(1)}M_{\lambda_0}^{(0)}X_0^\top X_0 w_0 ||_{\Sigma_1}^2 \right] + \sigma_0^2 \E\left[ || M_{\lambda_1}^{(1)} M_{\lambda_0}^{(0)} X_0^\top ||_{\Sigma_1, F}^2 \right]. $$

\begin{proof} First we notice the $\sigma_1$ terms of $R^S(\lambda_1)$ and $R^{TL}(\lambda_1 | \beta_0)$ exactly cancel. We are therefore left with
$$R^{S}(\lambda)- R^{TL}(\lambda) =  \E \left[ ||(M_\lambda^{(1)}X_1^\top X_1 - I)w_1||_{\Sigma_1}^2 \right]  -  \E \left[ || \lambda M_\lambda^{(1)}M_{\lambda_0}^{(0)}X_0^\top X_0 w_0 + \left(M_\lambda^{(1)} X_1^\top X_1 - I\right) w_1||_{\Sigma_1}^2 \right] $$
$$- \sigma_0^2 \lambda^2 \E \left[ || M_\lambda^{(1)}M_{\lambda_0}^{(0)}X_0^\top||_{\Sigma_1, F}^2 \right] $$

$$= -  \E \left[ || \lambda M_\lambda^{(1)}M_{\lambda_0}^{(0)}X_0^\top X_0 w_0 ||_{\Sigma_1}^2 \right] - 2 \lambda \E \left[ \braket{(M_\lambda^{(1)}X_1^\top X_1 - I)w_1, M_\lambda^{(1)}M_{\lambda_0}^{(0)}X_0^\top X_0 w_0 }_{\Sigma_1}\right]$$
$$-  \sigma_0^2\lambda^2 \E \left[ || M_\lambda^{(1)}M_{\lambda_0}^{(0)}X_0^\top||_{\Sigma_1, F}^2 \right] .$$

We note here that $(M_\lambda^{(1)} X_1^\top X_1 - I) = -\lambda M_\lambda^{(1)}$ and thus we have
$$\E \left[ \braket{(M_\lambda^{(1)}X_1^\top X_1 - I)w_1, M_\lambda^{(1)}M_{\lambda_0}^{(0)}X_0^\top X_0 w_0 }_{\Sigma_1}\right] = - \lambda \E \left[\braket{M_\lambda^{(1)} w_1, M_\lambda^{(1)}M_{\lambda_0}^{(0)}X_0^\top X_0 w_0 }_{\Sigma_1} \right]. $$

After some algebra and canceling the common $\lambda^2$ terms, we arrive at our desired inequality.
\end{proof}

\textbf{Corollary \ref{finite-ridgeless-boundary}} In the finite case, if $\Sigma_i = I$ and $\lambda_1 = \lambda_0 = 0$, and $X_i$ is taken to be Gaussian, then $R^{TL}(0) < R^{S}(0)$ if and only if:
$$ 2 \braket{w_0, w_1} > ||w_0||^2 + \sigma_0^2 \frac{p}{p-n_0-1}. $$

\begin{proof} We first note
$$\lim_{\lambda \searrow 0} M_{\lambda_1}^{(1)} = \lim_{\lambda \searrow 0} (X_1^\top X_1 + \lambda I)^{-1} = (X_1^\top X_1)^+ = I- X_1^+ X_1. $$
By conditioning on $X_0$ and taking expectation with respect to $X_1$, the projection $I - X_1^+ X_1$ introduces a common factor of $\frac{p-n_1}{p}$ to all terms involving $w_1$ or the $X_0$-dependent prior (due to the isotropy of $X_1$). Specifically, for any fixed vector $v$, $\E_{X_1}[||(I-X_1^+ X_1)v||^2] = \frac{p-n_1}{p}||v||^2$. We can thus eliminate the target projection effects from the expression. We are then left with:
$$2 \E \left[ \braket{  M_{\lambda_0}^{(0)}X_0^\top X_0 w_0 ,  w_1} \right] > \E \left[ || M_{\lambda_0}^{(0)}X_0^\top X_0 w_0 ||_{\Sigma_1}^2 \right] + \sigma_0^2 \E\left[ || M_{\lambda_0}^{(0)} X_0^\top ||_{\Sigma_1, F}^2 \right]. $$
We now note that
$$\lim_{\lambda_0 \searrow 0} M_{\lambda_0}^{(0)} X_0^\top = X_0^+, $$
and thus we have
$$2 \E \left[ \braket{  X_0^+ X_0 w_0 ,  w_1} \right] > \E \left[ || X_0^+ X_0 w_0 ||_{\Sigma_1}^2 \right] + \sigma_0^2 \E\left[ || X_0^+ ||_{\Sigma_1, F}^2 \right]. $$
We note here that since $\Sigma_0 = I$ we have $E \left[ X_0^+ X_0\right] = \frac{n_0}{p} I$. Additionally, $E\left[ ||X_0^+||_{F}^2\right] = \frac{n_0}{p-n_0-1}$. We can now take expectation to see
$$2 \frac{n_0}{p} \braket{  w_0 ,  w_1}  > \frac{n_0}{p} ||w_0 ||^2 + \sigma_0^2 \frac{n_0}{p-n_0-1}. $$
By multiplying by $\frac{p}{n_0}$ we arrive at the desired result.
\end{proof}

\subsection{Deterministic Equivalents and Asymptotics}

We will now examine asymptotics for the ridge phase transition identified above and use deterministic equivalents to understand the limiting phase transition. To establish the core deterministic equivalents, we must first let $\tau_i = \lambda_i / n_i$, and we define the following ridge resolvent:
Let $S_i := n_i^{-1} X_i^\top X_i$ and $\gamma_i := \lim_{p\to\infty} p/n_i$. Under standard Bai--Silverstein assumptions, the resolvent $(S_i+\tau_i I)^{-1}$ admits a deterministic equivalent
$$Q_i(\tau_i) = \Big(\tau_i I + \tilde{\delta}_i(\tau_i)\Sigma_i\Big)^{-1},$$
where the scalar pair $(\delta_i(\tau_i),\tilde{\delta}_i(\tau_i))$ is the unique positive solution to
$$\delta_i(\tau_i) = \frac{1}{n_i}\operatorname{Tr}\big(\Sigma_i Q_i(\tau_i)\big), \qquad \tilde{\delta}_i(\tau_i)=\frac{1}{1+\delta_i(\tau_i)}.$$

\textbf{Observation \ref{bai-silverstein-de}} Under standard Bai-Silverstein conditions (\cite{bai2010spectral}), as $p, n_i \rightarrow \infty$ with $p/n_i \rightarrow \gamma_i$:
$$n_1 M_{\lambda_1}^{(1)} = (S_1+\tau_1 I)^{-1} \asymp Q_1(\tau_1), $$
$$\lambda_1 M_{\lambda_1}^{(1)} \asymp \tau_1 Q_1(\tau_1),$$
and
$$n_0 M_{\lambda_0}^{(0)} X_0^\top X_0 M_{\lambda_0}^{(0)} \asymp Q_0(\tau_0) - \tau_0 Q_0(\tau_0)^2. $$

\begin{proof} These equivalences follow from standard results in \cite{bai2010spectral}. Specifically, under the Bai-Silverstein conditions, the empirical resolvent $(X_i^\top X_i/n_i + \lambda_i I)^{-1}$ admits the deterministic equivalent $n_i^{-1} Q_i(\tau_i) = n_i^{-1} (\tau_i I + \delta_i(\tau_i)\Sigma_i)^{-1}$, where $\delta_i(\tau_i)$ is the unique positive solution to the Silverstein fixed-point equation.  By dividing by $\lambda_1$ we arrive at the first deterministic equivalence.

On the sample side we know
$$M_{\lambda_0}^{(0)} X_0^\top = X_0^\top M_{\lambda_0}^{(0)} = n_0^{-1} X_0^\top \left( X_0^\top X_0 / n_0 + \lambda_0 I\right)^{-1}, $$
and the second deterministic equivalent follows.

Finally we use the following fact:
$$M_{\lambda_0}^{(0)} X_0^\top X_0 M_{\lambda_0}^{(0)} = M_{\lambda_0}^{(0)} \left( (X_0^\top X_0 +\lambda_0 I) - \lambda_0 I\right) M_{\lambda_0}^{(0)} = M_{\lambda_0}^{(0)} - \lambda_0 (M_{\lambda_0}^{(0)})^2. $$
From this it is easy to see the third deterministic equation holds.
\end{proof}

Finally, define $t(\tau_0, \tau_1)$ by: 
$$ \lim_{p \rightarrow \infty} p^{-1} \operatorname{Tr}\left( Q_1(\tau_1) \Sigma_1 Q_1(\tau_1) \left( Q_0(\tau_0) - \tau_0 Q_0(\tau_0)^2 \right)\right), $$
which we know exists under the Bai-Silverstein assumptions. By independence of $X_0$ and $X_1$, we may substitute these in to the results from Theorem~\ref{finite-benefit-analysis} to arrive at the following asymptotic decision criterion:

\textbf{Theorem \ref{asymptotic-benefit-analysis}} In the asymptotic limit, we gain benefit from transfer learning ($R^{TL}(\lambda_1) < R^S(\lambda_1)$) if and only if:
$$ 2 \braket{ Q_1(\tau_1) (I - \tau_0 Q_0(\tau_0)) w_0,  Q_1(\tau_1) w_1}_{\Sigma_1} >  ||  Q_1(\tau_1) (I - \tau_0 Q_0(\tau_0)) w_0 ||_{\Sigma_1}^2  +  \sigma_0^2 \gamma_0 t(\tau_0, \tau_1). $$

\begin{proof} This result follows directly from substituting the deterministic equivalents from Observation \ref{bai-silverstein-de} into the results from Theorem \ref{finite-benefit-analysis} combined with the independence of $X_1$ and $X_0$.
\end{proof}

\textbf{Corollary \ref{isotropic-asymptotic-benefit}} In the isotropic case where $\Sigma_0 = \Sigma_1 = I$, we (asymptotically) gain benefit from transfer learning ($R^{TL}(\lambda_1) < R^S(\lambda_1)$) if and only if:
$$ 2 \braket{w_0, w_1} >  (1-\tau_0 a_0) ||w_0||^2 + \sigma_0^2 \gamma_0 a_0, $$
where $a_0$ is the unique positive solution to
$$\tau_0 \gamma_0 a_0^2 + (\tau_0 + 1 - \gamma_0)a_0 - 1 = 0.$$

\begin{proof} We note that when $\Sigma_i = I$, the deterministic equivalent is scalar: $Q_i(\tau_i) = a_i I$ where $a_i$ is the unique positive solution to
$$\tau_i \gamma_i a_i^2 + (\tau_i + 1 - \gamma_i)a_i - 1 = 0.$$
We can substitute this in to the relevant quantities from Theorem \ref{asymptotic-benefit-analysis} to see:
$$ 2 \braket{ Q_1(\tau_1) (I - \tau_0 Q_0(\tau_0)) w_0,  Q_1(\tau_1) w_1}_{\Sigma_1}  = 2 a_1^2 (1-\tau_0 a_0)\braket{w_0,w_1} \text{ and} $$
$$||  Q_1(\tau_1) (I - \tau_0 Q_0(\tau_0)) w_0 ||_{\Sigma_1}^2 = a_1^2(1-\tau_0 a_0)^2||w_0||^2.$$

We now examine
$$t(\tau_0, \tau_1) = \lim_{p \rightarrow \infty} p^{-1} Tr \left( a_1^2 I (a_0 I - \tau_0 a_0^2 I) \right)$$
$$= a_1^2 Tr \left( a_0 I - \tau_0 a_0^2 I \right)/p$$
$$= a_1^2 (a_0 - \tau_0 a_0^2)$$
$$= a_1^2 a_0(1-\tau_0 a_0).$$
Substituting into Theorem~\ref{asymptotic-benefit-analysis} and canceling the common factor $a_1^2$, we obtain
$$2(1-\tau_0 a_0)\braket{w_0,w_1} > (1-\tau_0 a_0)^2 ||w_0||^2 + \sigma_0^2 \gamma_0 a_0(1-\tau_0 a_0).$$
Dividing by $(1-\tau_0 a_0)>0$ yields the stated condition.
\end{proof}

\textbf{Theorem \ref{optimal-source-ridge}} In the asymptotic setting with isotropic data and fixed source model overparameterization $\gamma_0$, for any fixed normalized alignment $\rho = \braket{w_0, w_1}/||w_0||^2$, there exists a unique source ridge penalty $\tau_0^*$ that maximizes the transfer benefit $\Delta R = R^S - R^{TL}$. Additionally, outside of a measure-zero set of parameters, the optimal $\tau_0^*$ for transfer learning differs from the optimal ridge penalty for Task~0 performance.

\begin{proof}
We seek to maximize the asymptotic risk benefit $\Delta R(\tau_0) = R^S - R^{TL}(\tau_0)$. Since $R^S$ is independent of $\tau_0$, this is equivalent to minimizing $R^{TL}(\tau_0)$. Using the isotropic deterministic equivalents derived in Corollary~\ref{isotropic-asymptotic-benefit}, the transfer risk is proportional to:
$$J(\tau_0) = (1-\tau_0 a_0)^2 ||w_0||^2 + \sigma_0^2 \gamma_0 a_0 (1-\tau_0 a_0) - 2 (1-\tau_0 a_0) \braket{w_0, w_1},$$
where $a_0$ is implicitly defined by $\tau_0$. Let $x(\tau_0) = 1 - \tau_0 a_0$. From the fixed-point equation $\tau_0 \gamma_0 a_0^2 + (\tau_0 + 1 - \gamma_0) a_0 - 1 = 0$, we can derive the bijection $x = \frac{a_0}{1+\gamma_0 a_0}$. Note that $a_0$ is strictly decreasing in $\tau_0$ (from $\infty$ to $0$), and $x$ is strictly increasing in $a_0$ (mapping $(0, \infty)$ to $(0, \gamma_0^{-1})$). Thus, optimizing with respect to $\tau_0$ is equivalent to optimizing with respect to the shrinkage factor $x$.
Substituting $a_0 = \frac{x}{1-\gamma_0 x}$ into the objective:
$$J(x) = x^2 ||w_0||^2 + \frac{\sigma_0^2 \gamma_0 x^2}{1-\gamma_0 x} - 2 x \braket{w_0, w_1}.$$
The derivative with respect to $x$ is:
$$J'(x) = 2x ||w_0||^2 + \sigma_0^2 \gamma_0 \frac{2x(1-\gamma_0 x) + \gamma_0 x^2}{(1-\gamma_0 x)^2} - 2 \braket{w_0, w_1} = 2x ||w_0||^2 + \sigma_0^2 \gamma_0 \frac{2x - \gamma_0 x^2}{(1-\gamma_0 x)^2} - 2 \braket{w_0, w_1}.$$
Let $G(x) = x ||w_0||^2 + \frac{\sigma_0^2 \gamma_0}{2} \frac{x(2-\gamma_0 x)}{(1-\gamma_0 x)^2}$. For $x \in (0, \gamma_0^{-1})$, $G(x)$ is strictly increasing. The optimal shrinkage $x^*$ is the unique solution to $G(x^*) = \braket{w_0, w_1}$.
Define the source-optimal shrinkage $x_S^*$ as the minimizer of the source risk (which corresponds to setting $w_1 = w_0$ in the transfer risk). Thus $x_S^*$ satisfies $G(x_S^*) = ||w_0||^2$.
Comparing the two conditions:
$$G(x^*) = \braket{w_0, w_1} \quad \text{and} \quad G(x_S^*) = ||w_0||^2.$$
Since $G$ is strictly increasing/injective, $x^* = x_S^*$ if and only if $\braket{w_0, w_1} = ||w_0||^2$. Thus, outside the measure-zero set where alignment exactly equals source signal power, $x^* \neq x_S^*$, implying $\tau_0^* \neq \tau_S^*$.
\end{proof}

\textbf{Corollary \ref{snr-threshold-cor}} Let $\tau_0^*$ be the transfer-optimal regularization penalty and $\tau_S^*$ be the source-optimal regularization penalty.
\begin{itemize}
    \item If the tasks are imperfectly aligned ($0<\rho<1$), then $\tau_0^* > \tau_S^*$ (transfer requires stronger regularization).
    \item If the tasks are super-aligned ($\rho>1$), then $\tau_0^* < \tau_S^*$ (transfer requires weaker regularization).
\end{itemize}
This phase transition depends solely on task alignment and holds for all noise levels $\sigma_0 > 0$.

\begin{proof}
Recall from the proof of Theorem~\ref{optimal-source-ridge} that the optimal shrinkage factors $x^*$ (transfer) and $x_S^*$ (source) satisfy:
$$G(x^*) = \braket{w_0, w_1} \quad \text{and} \quad G(x_S^*) = ||w_0||^2,$$
where $G(x)$ is a strictly increasing function.
Recall also that $x(\tau_0) = 1 - \tau_0 a_0$ is a strictly decreasing function of the regularization $\tau_0$.
\begin{itemize}
    \item Case 1: Imperfect alignment ($\braket{w_0, w_1} < ||w_0||^2$). Then $G(x^*) < G(x_S^*)$, which implies $x^* < x_S^*$. Since $x$ decreases with $\tau_0$, this implies $\tau_0^* > \tau_S^*$.
    \item Case 2: Super-alignment ($\braket{w_0, w_1} > ||w_0||^2$). Then $G(x^*) > G(x_S^*)$, which implies $x^* > x_S^*$. Since $x$ decreases with $\tau_0$, this implies $\tau_0^* < \tau_S^*$.
\end{itemize}
\end{proof}

%%%%%%%%%%%%%%%%%%%%%%%%%%%%%%%%%%%%%%%%%%%%%%%%%%%%%%%%%%%%%%%%%%%%%%%%%%%%%%%
%%%%%%%%%%%%%%%%%%%%%%%%%%%%%%%%%%%%%%%%%%%%%%%%%%%%%%%%%%%%%%%%%%%%%%%%%%%%%%%

\end{document}

%% file: math_commands.tex
\usepackage{amsmath,amsfonts,bm}

\def\eqref#1{equation~\ref{#1}}
\def\1{\bm{1}}

\DeclareMathAlphabet{\mathsfit}{\encodingdefault}{\sfdefault}{m}{sl}
\SetMathAlphabet{\mathsfit}{bold}{\encodingdefault}{\sfdefault}{bx}{n}

\newcommand{\E}{\mathbb{E}}